%% file: main.tex
\pdfoutput=1

\documentclass[11pt]{article}

\usepackage[]{EMNLP2022}

\usepackage{times}
\usepackage{latexsym}

\usepackage[T1]{fontenc}

\usepackage[utf8]{inputenc}

\usepackage{microtype}

\usepackage{inconsolata}
\usepackage{booktabs}
\usepackage{xcolor}         %
\usepackage{graphicx}
\usepackage{xspace}
\usepackage{todonotes}
\usepackage{multirow}
\usepackage{wrapfig}
\usepackage{caption}
\usepackage{subcaption}
\usepackage{tabularx}
\usepackage{amsmath}
\usepackage{bbm}
\usepackage{bm}
\usepackage{mathtools}
\usepackage{lipsum}
\usepackage[ruled,vlined]{algorithm2e}

\usepackage{pifont}%

\newcommand{\eat}[1]{}

\newcommand{\method}{EfficientVLM\xspace}
\newcommand{\param}{\bm{\theta}}
\newcommand{\lzero}{$L_0$\ }
\newcommand{\mask}{\mathbf{z}}

\def\Vec#1{{\boldsymbol{#1}}}

\title{EfficientVLM: Fast and Accurate Vision-Language Models via \\ Knowledge Distillation and Modal-adaptive Pruning}

\author{Tiannan Wang\thanks{\ Equal contribution, work done during internship at Bytedance AI Lab} \\
  Beihang University \\
  \texttt{tiannanwang@buaa.edu.cn} \\ \And
  Wangchunshu Zhou*\\
  ETH Zurich \\
  \texttt{wangchunshu.zhou@inf.ethz.ch} \\ \AND
  Yan Zeng \\
  Bytedance AI Lab \\
  \texttt{zengyan.yanne@bytedance.com} \\ \And
   Xinsong Zhang \\
  Bytedance AI Lab \\
  \texttt{zhangxinsong.0320@bytedance.com} \\
  }

\begin{document}
\maketitle

\input{sections/0_abstract}
\input{sections/1_introduction}

\input{sections/2_method}

\input{sections/3_experiments}

\input{sections/4_conclusion}

\section*{Limitations}

\method is applied on X-VLM. However, there are also many recent fully Transformer VLMs achieving comparable or better performance. Therefore, applying our \textit{distilling then pruning} framework on other state-of-the-art VLMs can be interesting. Also, we do not apply quantization or matrix decomposition, which are also prevalent model compression techniques.

\section*{Ethics Statement}

Our method is used to compress VLMs. Therefore, ethic considerations of VLMs generally apply to our method. We encourage users to assess potential biases before deploying \method.

\bibliography{custom}
\bibliographystyle{acl_natbib}

\clearpage
\appendix
\input{sections/5_appendix}

\end{document}

%% file: sections/0_abstract.tex
\begin{abstract}

Pre-trained vision-language models (VLMs) have achieved impressive results in a range of vision-language tasks. 
However, popular VLMs usually consist of hundreds of millions of parameters which brings challenges for fine-tuning and deployment in real-world applications due to space, memory, and latency constraints. 
In this work, we introduce a \textit{distilling then pruning} framework to compress large vision-language models into smaller, faster, and more accurate ones. 
We first shrink the size of a pre-trained large VLM and apply knowledge distillation in the vision-language pre-training stage to obtain a task-agnostic compact VLM. 
Then we propose a modal-adaptive pruning algorithm to automatically infer the importance of vision and language modalities for different downstream tasks and adaptively remove redundant structures and neurons in different encoders with controllable target sparsity. 

We apply our framework to train \method, a fast and accurate vision-language model consisting of 6 vision layers, 3 text layers, and 3 cross-modal fusion layers, accounting for only 93 million parameters in total, which is 44.3\% of the teacher model. \method retains 98.4\% performance of the teacher model and accelerates its inference speed by $2.2\times$. \method achieves a large absolute improvement over previous SoTA efficient VLMs of similar sizes by a large margin on various vision-language tasks, including VQAv2 (+4.9\%), NLVR2 (+5.6\%), ITR (R@1 on TR +17.2\%, on IR + 15.6\% ) and COCO caption generation (CIDEr +6.5), demonstrating a large potential on training lightweight VLMs. \footnote{Our code and pretrained checkpoints are available at \url{https://github.com/swaggy-TN/EfficientVLM}.}

\end{abstract}

%% file: sections/1_introduction.tex
\section{Introduction}
Inspired by the success of large pre-trained language models \cite{devlin2018bert,radford2018improving} in the field of natural language processing (NLP), recent studies~\cite{su2019vl,li2020unicoder,alec2021clip,kim2021vilt,li2020unimo} in vision-language pretraining (VLP) have advanced the state-of-the-art on various vision-language tasks such as image captioning, visual question answering, and image-text retrieval. 

However, in both NLP and vision-language domains, large Transformer-based pre-trained models often consist of hundreds of millions, if not billions, of parameters, bringing various practical challenges for deployment.
As summarized in~\citet{schwartz2020green} and \citet{xu2021survey}, large pre-trained models require large amounts of space (in terms of GPU memory and disk storage) and heavy computing for fine-tuning and inference, which is both costly and may lead to negative environmental impact. Furthermore, large models inevitably lead to low latency, which poses a challenge for the production environment. 

Recent literature revealed that BERT~\citep{devlin2018bert}, a popular Transformer-based pre-trained language model, can be effectively compressed and accelerated via knowledge distillation~\citep{sanh2019distilbert,jiao2019tinybert,xu2020bert,wang2020minilm}. However, only a few prior works investigated building efficient VLMs. For instance, \citet{wang2020minivlm} introduced MiniVLM which combines a lighter object detector with a compressed BERT~\citep{wang2020minilm}.  \citet{fang2021compressing} further proposed DistilVLM, which uses knowledge distillation to pre-training a compact VLM with the guidance from a large pre-trained VLM. However, their approach is limited to object-feature-based VLMs. As such, the vision feature extractor cannot be distilled together with the Transformer model in an end-to-end manner, which limits the potential of knowledge distillation. As a result, existing compact VLMs are generally falling short compared to regular-size VLMs.

In this work, we investigate strategies for VLM compression and introduce a \textit{distilling then pruning} framework for compressing fully Transformer-based VLMs.  Specifically, in the first stage, we use knowledge distillation for task-agnostic compression of a pre-trained VLM by aligning the logits, attention distribution, and hidden representations between the student model and the teacher model. This results in a  \textbf{task-agnostic} compact VLM that achieves competitive results on many downstream vision-language tasks by simply fine-tuning. The general distillation stage reduces the size of all modules (i.e., vision encoder, text encoder, cross-modal encoder) equally so that the compressed model can be versatile to different downstream tasks. However, our preliminary study, which is described in detail in section~\ref{sec:modal adaptive}, shows that not all modules are created equal in a VLM and their importance drastically varies on different downstream vision-language tasks requiring different level of understanding on either vision and text modalities. This 
indicates that compressing a VLM requires modal- and \textbf{task-specific} designs. Therefore, in the second stage, we propose to prune the compact VLM when fine-tuning on different downstream tasks to flexibly adjust the model size/latency according to modal importance. Concretely, we propose a modal-adaptive pruning strategy that regularizes the model with a differentiable approximation to the $L_0$-norm regularization~\citep{louizos2017learning} to automatically infer the importance of vision and language modalities with controllable target sparsity. In this way, our method can adaptively prune different modules in the VLM in the fine-tuning stage according to the relative importance of vision-language modalities on different downstream tasks.

We apply our framework to compress X-VLM~\citep{zeng2021multi}, a recent Transformer-based VLM and train \method, a fast and accurate vision-language model. \method consists of 6 vision layers, 3 text layers, and 3 cross-modal fusion layers, accounting for only 93 million parameters in total, which is 44.3\% of the X-VLM model. \method recovers 98.4\% performance of X-VLM and accelerates its inference speed by $2.2\times$. 
Experimental results show that despite being trained with fewer image-text pairs, \method achieves a large absolute improvement over DistilVLM, the previous best-performing efficient VLM with similar size and inference speed, on various vision-language tasks, including VQAv2~\citep{goyal2017making} (+6.7\%), NLVR2~\citep{suhr2018corpus} (+7.8\%), ITR-COCO~\citep{lin2014microsoft} (R@1 on TR +19.9\%, R@1 on IR + 15.6\% ) and COCO caption generation~\citep{chen2015microsoft} (CIDEr +6.5), demonstrating a large potential on training lightweight VLMs.

To the best of our knowledge, our work is the first attempt to (1) compress a fully Transformer-based vision-language model, and (2) combine knowledge distillation with (modal-adaptive) pruning for vision-language model compression.

\section{Related Work}

\paragraph{Vision-Language Pre-training}
The existing work on vision language pre-training typically falls into two categories. Most methods rely on object detection~\cite{tan2019lxmert, lu2019vilbert, li2019visualbert, su2019vl, li2020unicoder, chen2020uniter, li2020oscar, gan2020large, li2020unimo, xu2021e2e, liu2021kd, li2022uni}, where an image is represented by dozens of object-centric features. However, the object detection process requires high-resolution images as model input and is very time-consuming. Moreover, most works under this category utilize pre-trained object detectors~\cite{ren2015faster, anderson2018bottom}, and do not optimize the model in an end-to-end manner, yielding sub-optimal performance.  Therefore, recent works turn to encoding images by convolutional network~\cite{jiang2020defense, huang2020pixel, huang2021seeing, wang2022unifying} or vision transformer~\cite{kim2021vilt, li2021align}, largely improving the inference speed. Nevertheless, some recent work~\cite{zhang2021vinvl,zeng2021multi} shows that understanding fine-grained vision language alignments (e.g. object-level) is critical for some downstream tasks such as visual reasoning and visual grounding. 

\begin{figure*}
    \centering
    \includegraphics[width=\textwidth]{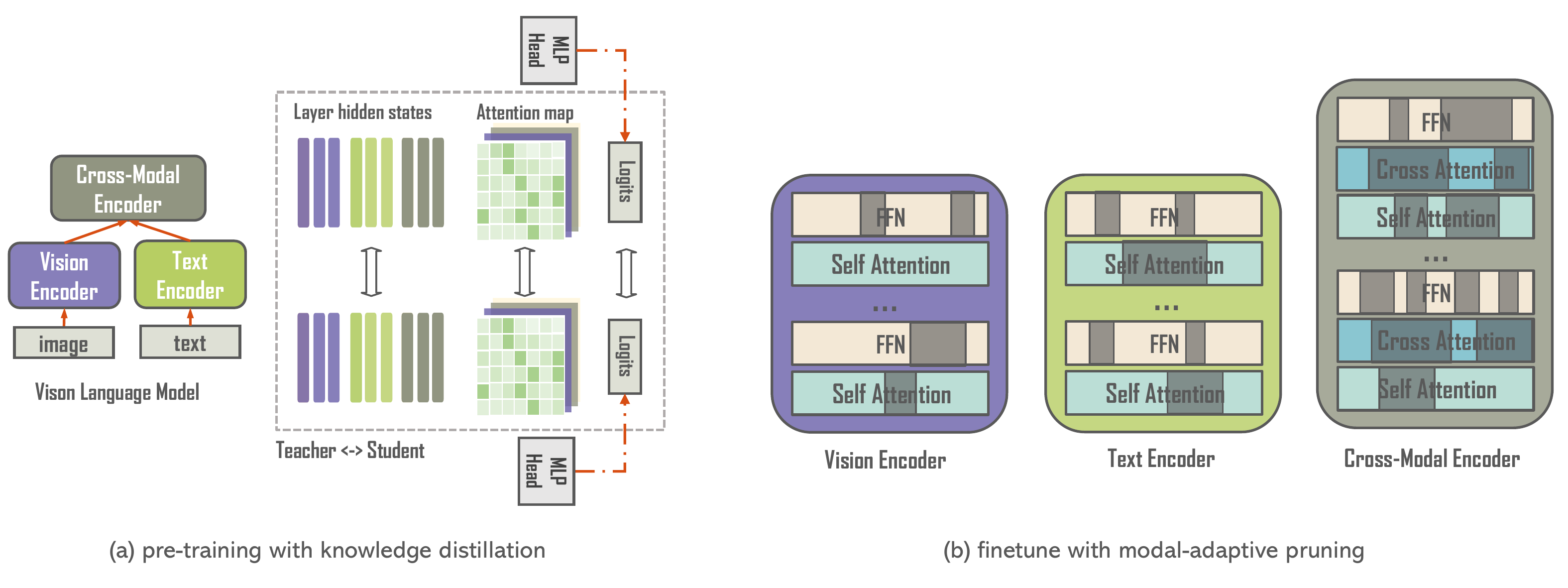}
    
    \caption{The \textit{distilling then pruning} framework for training \method. In the pre-training stage, we apply knowledge distillation with a pre-trained X-VLM model as the teacher. During fine-tuning, we use a modal-adaptive pruning method to adaptively prune encoders of different modalities.}
    \label{fig:model}
    \vspace{-3mm}
\end{figure*}

\paragraph{Pre-trained Model Compression}
Prior work has shown that BERT~\citep{devlin2018bert}, a popular encoder-only pre-trained Transformer~\citep{vaswani2017attention}, can be effectively compressed and accelerated. As summarized in \citet{xu2021survey}, popular BERT compression techniques include knowledge distillation~\citep{hinton2015distilling,sanh2019distilbert,sun2019patient,jiao2019tinybert,wang2020minilm,zhou2022bert,xu2021beyond} which trains a compact student network to mimic the behavior of the original teacher model, pruning~\citep{lecun1989optimal,michel2019sixteen,gordon2020compressing,sanh2020movement,lagunas2021block,wang2019structured,xia2022structured} which prunes redundant neurons or structures in the original model, module replacing~\citep{xu2020bert} which train compact successor sub-modules to replace that in the original model, and quantization~\citep{shen2020q,zafrir2019q8bert} that compresses a neural network by reducing the number of bits used to represent its parameters. On the other hand, a number of work also investigated efficient inference with BERT-like models with early exit~\citep{teerapittayanon2016branchynet,xin2020deebert,liu2020fastbert,schwartz2020right,zhou2020bert} or adaptive computation time~\citep{graves2016adaptive,eyzaguirre2021dact}.

In contrast, only a few prior work investigated methods to compress a pre-trained vision-language model. \citet{fang2021compressing} explored distilling a pre-trained vision-language model into a more compact student model and proposed a teacher adaptation method that aligns object feature proposal. However, their approach is limited to the use of an object detection based vision-language model, which makes end-to-end distillation infeasible and results in unsatisfactory performance compared to recent state-of-the-art. \citet{wang2021distilled} explored distilling a vision-language model with a cross-modal fusion module to a dual-encoder model for efficient retrieval. Moreover, \citet{gan2021playing} explored the lottery ticket hypothesis~\citep{frankle2018lottery} in vision-language models and find that sparse winning tickets exist in pre-trained VLMs. However, the process of finding and re-training winning tickets are less efficient compared to other compression 
methods.

%% file: sections/2_method.tex
\section{\method}
In this section, we present \method, a fast and accurate vision-language model trained with our \textit{distilling then pruning} framework. We choose X-VLM~\citep{zeng2021multi}, one of the state-of-the-art vision-language model, as the teacher model. \footnote{In practice, our proposed method suits for any VLMs that equipped with modal-specific module such as VLMo~\citep{bao2022vlmo} or ALBEF~\citep{li2021align}.}

\subsection{Model Overview}

\method is a compressed version of X-VLM, a fully Transformer-based VLM. X-VLM has the same architecture as ALBEF~\citep{li2021align}, which consists of an image encoder, a text encoder, and a cross-modal encoder. The image encoder contains 12 transformer layers, while the text encoder and the cross-modal encoder consist of 6 transformer layers respectively. The cross-modal encoder fuses the vision features with the text features through cross-attention at each layer.  
\method shrinks the size of X-VLM by half, thus consisting of 6 vision layers, 3 text layers, and 3 cross-modal layers, accounting for only 92 million parameters in total, which is 43.6\% of the X-VLM model.

The teacher model is optimized by: 1) aligning the texts and visual concepts, where the alignments are in multi-granularity using a contrastive loss $\mathcal{L}_\text{ITC}$, a matching loss $\mathcal{L}_\text{ITM}$, and a masked language modeling loss $\mathcal{L}_\text{MLM}$; 2) in the meantime locating visual concepts in the image given the corresponding texts by bounding box prediction loss $\mathcal{L}_\text{BBOX}$. Overall, the vision language pre-training loss is: 
\begin{align}
    \mathcal{L}_{\text{VLP}} =  \mathcal{L}_\text{ITC} + \mathcal{L}_\text{ITM} + \mathcal{L}_\text{MLM} + \mathcal{L}_\text{BBOX}
\end{align}

\subsection{Pre-training with Knowledge Distillation}
We initialize \method with a pre-trained X-VLM and shrink its size by half by only retaining the even-numbered layers. Then we pre-train \method on image-text pairs with both the original vision-language pre-training objectives of X-VLM and knowledge distillation objective with the pre-trained X-VLM as the teacher model. The knowledge distillation objective consists of attention distillation, hidden states distillation, and logits distillation.

\paragraph{Attention Distillation}
Prior work~\citep{jiao2019tinybert} on BERT distillation have shown the effectiveness of transferring the latent knowledge in self-attention matrices:

\begin{equation}
    \mathbf{A}=\operatorname{softmax(}{\mathbf{Q}\cdot \mathbf{K}}/{\sqrt{d_k}}\text{)}.
\end{equation}
where $\mathbf{Q}$ and $\mathbf{K}$ denotes the query and key matrix in the attention layer of a transformer block. $\mathbf{d_k}$ is the dimension of the key matrix as a scaling factor.
We formulate attention distillation loss by minimizing the mean square error between the self-attention matrices of the teacher and the student:
\begin{equation}
\label{eq:att_loss}
\mathcal{L}_{\text{attn}} = \frac{1}{h}\sum\nolimits^{L}_{j=1}\sum\nolimits^{h}_{i=1} \operatorname{MSE}(\bm{A}_{i,j}^{S}, \bm{A}_{i,2j}^{T})
\end{equation}
where $L$ denotes the number of layer in each encoder of the student, $h$ is the number of attention heads, $\bm{A}_i$ refers to the normalized attention matrix corresponding to the $i$-th head  in $j$-th layer of the student and in $2j$-th layer of the teacher. 
The attention matrix is in shape of $\bm{A} \in \mathbbm{R}^{l\times p}$. $l$ and $p$ are the length of query and key, respectively\footnote{In the cross-attention module of cross-modal encoder, $p$ represents the length of patch sequence of vision encoder otherwise $l$ and $p$ are equal}.

\paragraph{Hidden States Distillation}
Following Transformer distillation in TinyBERT~\cite{jiao2019tinybert}, we also adopt the hidden states distillation to better utilize the information from the teacher model. The loss function is defined as follows:
\begin{equation}
    \mathcal{L}_{\text{hid}} = \sum\nolimits^{L}_{i=1}\text{MSE}(\bm{H}_{i}^{S}, \bm{H}_{2i}^{T}), 
\end{equation}
 $\bm{H}^{S} \in \mathbbm{R}^{l\times d'}$ and $\bm{H}^{T} \in \mathbbm{R}^{l \times d}$ refer to the hidden states of student and teacher networks in the corresponding layer.

\paragraph{Logits Distillation}
In addition to imitating the behaviors of intermediate layers, we also use the knowledge distillation to fit the predictions of teacher model as in~\cite{hinton2015distilling}. We adopt KL divergence as the optimization objective:

\paragraph{Pre-training}
We formulate the final loss by combing the original vision-language pre-training loss with general distillation loss.
\begin{align*}
    &\mathcal{L}_\text{KD} = \alpha\mathcal{L}_\text{attn} + \beta\mathcal{L}_\text{hid} + \gamma\mathcal{L}_\text{logits} \\
     &\mathcal{L}_\text{pretrain} = \lambda\mathcal{L}_\text{VLP} + (1-\lambda)\mathcal{L}_\text{KD}
\end{align*}
where $\alpha$, $\beta$, $\gamma$ and $\lambda$ are the weights of the loss terms. We only adjust the weights to scale the losses to similar values so that the optimization process can perform more robust.

\subsection{Fine-tuning with Pruning}
To flexibly adjust the efficiency-performance trade-off of \method on different downstream tasks according to varying resource constraints, we propose a modal-adaptive pruning method to further compress \method to a desired size in the fine-tuning stage. 

\begin{figure}
\centering
\begin{minipage}{.25\textwidth}
  \centering
  \includegraphics[width=\textwidth]{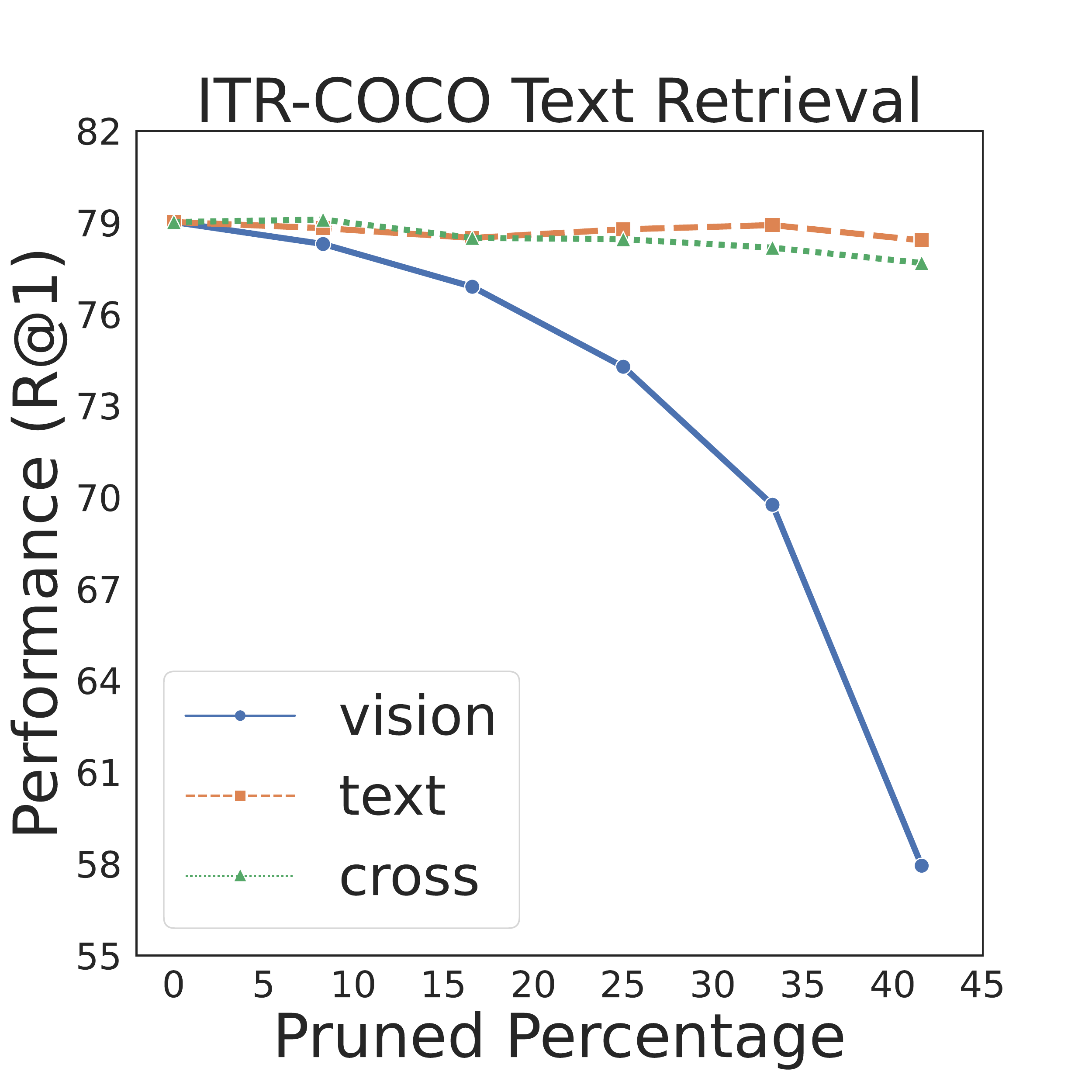}
\end{minipage}%
\begin{minipage}{.25\textwidth}
  \centering
  \includegraphics[width=\textwidth]{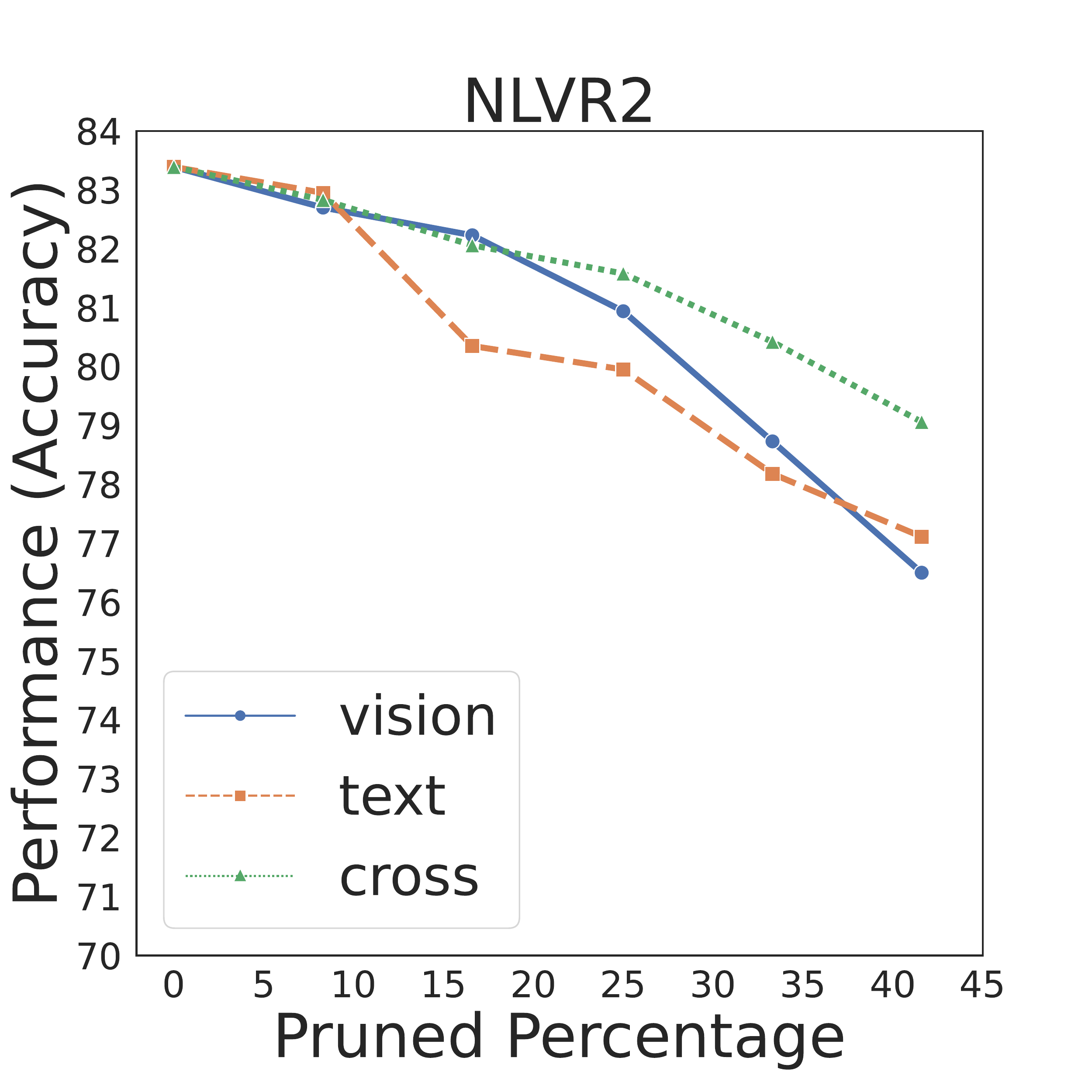}
\end{minipage}
\caption{Empirical study of modal-encoders importance on NLVR2 and ITR-COCO tasks.} 
\vspace{-0.4cm}
\label{fig:empirical}
\end{figure}

\paragraph{Are All Modalities Created Equal in VLMs?}
Unlike prior work~\citep{lagunas2021block} on BERT pruning where there is only one Transformer encoder, pruning VLMs are more challenging because the importance of vision and language clues may not be equally important~\citep{cao2020behind}. This is also verified by our preliminary experiments where we prune 40\% attention heads in each encoder and find that the performance drops drastically, which is contrary to prior findings on pruning BERT~\citep{michel2019sixteen}.

To this end, we conduct an empirical study to investigate whether encoders for vision / language modalities have similar importance across different vision-language tasks.
We prune each encoder in a fine-tuned teacher model at one time while leaving other encoders untouched. From Figure \ref{fig:empirical}, we observe that: (1) the encoders of different modalities have different sensitivity with respect to head pruning, and (2) the difference in sensitivity varies on different downstream tasks. Specifically, on ITR-COCO task, pruning 40\% heads in text encoder and cross-modal encoder does not significantly impact performance while pruning vision encoder causes a large performance drop. However, the results on NLVR2 show that text encoder is as important as image encoder in this task while cross-modal encoder are not very sensitive to head pruning. These results suggest that encoders of different modalities are not created equal in a vision-language model, motivating us to explore modal-specific pruning methods for VLMs.

\paragraph{Modal-adaptive pruning}
\label{sec:modal adaptive}

A naive way to achieve modal-specific pruning is to manually adjust the pruning percentage of different encoders based on the prior observation. Specifically, we consider a baseline that prune 30\% parameters out of each encoder as the baseline. Then for ITR-COCO, we prune 10\% parameters in the vision encoder while pruning 40\% parameters in the text and the cross-modal encoder. For NLVR2, we set this percentage to 10\%, 10\%, and 60\% for image, text, and cross-modal encoders, respectively. These percentages are heuristically adjusted according to the previous findings and the empirical performance. Moreover, the relative sparsity is set to ensure the overall sparsity of the model is similar. 

\begin{table}[h]
    \centering
    \resizebox{\linewidth}{!}{
    \begin{tabular}{c|ccc|ccc|cc}
    \toprule
        sparsity & \multicolumn{3}{c|}{Text Retrieval} & \multicolumn{3}{c|}{Image Retrieval} &\multicolumn{2}{c}{NLVR2} \\
         & R@1 & R@5 & R@10 &  R@1 & R@5 & R@10 & val & test \\
         \midrule
        .3/.3/.3 & 76.4 & 93.4 & 96.8 & 58.6 & 83.2 & 90.0 & 78.9 & 77.9 \\
        .1/.4/.4 & 78.1 & 94.2 & 97.1 & 60.2 & 84.1 & 90.5 & - & - \\
        .1/.1/.6 & - & - & - & - & - & - & 80.9 & 80.9 \\
        \bottomrule
    \end{tabular}}
    \caption{Modal-specific pruning results on NLVR2 and ITR-COCO. All models are trained with pruning and knowledge distillation.}
    \label{tab:observation}
    \vspace{-2mm}    
\end{table}

The results are shown in Table~\ref{tab:ablation}. We find that manually specifying sparsity levels for different encoders according to their ``importance'' leads to substantial improvements, demonstrating the effectiveness of modal-specific pruning. However, manually determining the sparsity for different encoders could be laborious and sub-optimal. Therefore, we propose \textbf{modal-adaptive pruning}, an end-to-end pruning algorithm using a differentiable approximation of $L_0$ regularization \citep{louizos2017learning} to automatically
infer the importance of vision and language modalities and adaptively remove redundant structures and neurons in different encoders with controllable target sparsity.

Consider a given neural network model $f(\cdot; \param)$ parameterized by $\param = \{ \theta_j \}_{j=1}^{n}$,
where each $\theta_j$ represents an individual parameter weight or a block of weights (e.g. a column of a weight matrix) and $n$ denotes the number of blocks.
A pruning strategy of the model can be parameterized by introducing additional binary variables $\mask = \{ z_j \}_{j=1}^{n}$ such that $z_j \in \{0, 1\}$ and
\begin{align*}
\tilde{\param} = \param\odot \mask \quad\qquad \forall j \ \ \tilde{\theta}_j = \theta_j \, z_j .
\end{align*}
Here $\tilde{\param} = \{ \tilde{\theta}_j \}$ denotes the set of model parameters after pruning and its \lzero ~norm, $\|\tilde{\param}\|_0 = \sum_{j=1}^n z_j$, measures the effective size of the pruned model.
The optimization during training can be formulated as minimizing the objective below
\begin{align} 
  \mathbbm{E}_{\textbf{z}} \left[ \, \frac{1}{D} \sum_{i=1}^D \mathcal{L}\left(\mathbf{x}_i, \mathbf{y}_i; \tilde{\param} \right) + \lambda \|\tilde{\param}\|_0 \, \right]
  \label{eq:pruning formulation}
\end{align}
where $\{\mathbf{x}_i, \mathbf{y}_i\}_{i=1}^D$ are training examples, $\mathcal{L}$ is the training loss function and $\lambda > 0$ is a constant hyper-parameter.
During training, the masking variables $\mask$ are learned as real numbers in range [0, 1] while during inference all the variable that below a threshold are set to 0 so that our pruned model can achieve the expected sparsity.
See Appendix~\ref{app:optimization} for more details.

We also adopt knowledge distillation at fine-tuning with pruning stage to help the student model better preserving capacity on downstream tasks. The final training objective is as follows:
\begin{equation}
    \mathcal{L}_\text{ft} = \lambda\mathcal{L}_\text{VL} + (1-\lambda)\mathcal{L}_\text{KD} + \mathcal{L}_\text{Lgr}
\end{equation}
where $\mathcal{L}_\text{VL}$ represents the task-specific fine-tuning loss brought by the re-parameterized student model, the $\mathcal{L}_\text{KD}$ is the task-specific knowledge distillation loss and $\mathcal{L}_\text{Lgr}$ infers to the lagrangian loss.

%% file: sections/3_experiments.tex
\section{Experiments}

\subsection{Baselines}

\begin{table*}[!]
    \small
    \centering
    \resizebox{\textwidth}{!}{
    \begin{tabular}{l|c|c|ccc|ccc}
    \toprule
        Method & Input Length & End-to-End &\multicolumn{3}{c|}{Vision Module} & \multicolumn{3}{c}{Text(and)Fusion Module}  \\
         & image/text & Time(ms) &Para(M) & Time(ms) & FLOPs(B) & Para(M) & Time(ms) & FLOPs(B) \\
    \midrule
        X-VLM$_{clip}$ & $196/35$ & $17.8$ &$86.1$ & $9.0$ & $18.9$ & $123$ & $8.8$($14.2$) & $4.2$ \\
        \ - CPU & - & $395.5$ & - &$355.1$ & - & - & $40.4$($56.8$) & - \\
        OSCAR$_{B}$ & $50/35$ &$135.2$ &$63.8$ &  $121.9$ & $767.0$  & $109$ & $13.3$ & $8.2$ \\
        \ - CPU  & - & $12347.1$ & - &$12300$ & - & - & $47.1$ & - \\
    \midrule
        MiniVLM & $50/35$ & $23.6$ &$7.5$ & $12.2$ & $4.4$ & $34.5$ & $11.4$ & $2.3$\\
        \ - CPU & - & $418.2$ & - & $393.9$ & - &- & $24.3$ & - \\
        ViLT  & $200/40$ & $21$ & $2.4$ &  $0.7$ & $0.6 $  & $109$ & $20.3$ & $22.8$ \\
        \ - CPU  & - & $69.1$ & - & $0.8$ & - & - & $68.3$ & - \\
    \midrule
        \method &$196/35$ & $9.7$ &$42.0$ & $5.0$ & $8.3$ & $50.3$ & $8.5$ & $1.3$\\ 
        \ - CPU & - & 180.8 &- & $171.5$ & - & - & $17.1$ & - \\ 
    \bottomrule
    \end{tabular}}
    \caption{Model size and actual inference time for visual feature extractor and vision-language fusion model of compared models. DistilVLM is of the same size and speed of MiniVLM. Actual Inference time is reported on both GPU and CPU.}
    
    \label{tab:infertime}
    \vspace{-2mm}    
\end{table*}

We mainly compare \method with two baselines: MiniVLM \citep{wang2020minivlm}, a compact VLM consists of a lightweight object detection model and a compact Transformers-based vision-language encoder, which is initialized by MiniLM~\citep{wang2020minilm}, a compressed pre-trained language model; and DistillVLM\citep{fang2021compressing}, which adopts the same model architecture with MiniVLM and apply knowledge distillation for further boosting model's performance. For reference, we also include the performance of DistilDualEnc~\citep{wang2021distilled}, ViLT ~\citep{kim2021vilt} and X-VLM$_\text{small}$ in our comparison. DistillDualEnc is a dual-encoder VLM distilled from a fusion-based VLM, ViLT is a single-stream VLM that feeds vision features without using region features nor deep convolutional visual embedders and X-VLM$_\text{small}$ use the same initialization as EfficientVLM but trained without knowledge distillation or pruning.

To make our comparison clearer, we present the size and inference speed of compared models in Table~\ref{tab:infertime}. We test model inference time on both GPU and CPU devices which are Nvidia Tesla V100 GPU and Intel(R) Xeon(R) Platinum 8260 CPU @2.40GHz, respectively. Since the number of FLOPs is affected by the input sequence length, we show the input image token length and average text length of each model in their settings in the table.
We can see that despite the fully Transformer-based visual feature extractor being heavier on model size, it consumes much less time during inference comparing to MiniVLM. As for the Transformer-based text/fusion module, \method is slightly larger than MiniVLM and DistilVLM while much faster thanks to the parallel nature of image and text encoders in its architecture. Despite the extremely efficient vision module of ViLT, it consume more time because of its heavy text and fusion encoder.
Specifically, when comparing with their corresponding teacher model, DistilVLM only reduces the inference time of the Transformer encoder by around 15\% on GPU, while \method achieves a speed-up ratio of 1.9$\times$ on GPU and 2.2$\times$ on CPU.

\begin{table*}[h]
    \centering
    \resizebox{\textwidth}{!}{
    \begin{tabular}{l|ccc|ccc|cc|cc|cccc}
    \toprule
        Method & \multicolumn{3}{c|}{ITR-TR} & \multicolumn{3}{c|}{ITR-IR} &\multicolumn{2}{c|}{NLVR2} &\multicolumn{2}{c|}{VQA 2.0} &\multicolumn{4}{c}{COCO-Caption}\\
         & R@1 & R@5 & R@10 &  R@1 & R@5 & R@10 & val & test & test-dev & test-std & B@4 & M & C & S\\
         \midrule
    
    X-VLM$_\text{clip}$ & 79.0 & 94.5 & 97.9 & 61.5 & 84.6 & 90.8 & 83.15 & 83.48 & 76.92 & 77.02 & 39.4 & 30.5 & 131.0 & 23.6 \\
    --98\% & 77.4 & 92.6 & 95.9 & 60.3 & 82.9 & 89.0 & 81.49 & 81.81 & 75.38 & 75.48 & 38.6 & 29.9 & 128.4 & 23.1 \\
    OSCAR$_\text{B}$ & 70.0 & 91.1 & 95.5 & 54.0 & 80.8 & 88.5 & 78.07 & 78.36 & 73.4 & 73.2 & 36.5 & 30.3 & 123.7 & 23.1\\
    --98\% & 68.6 & 89.3 & 93.6 & 52.9 & 79.2 & 86.7 & 76.51 & 76.79 & 71.93 & 71.74 & 35.8 & 29.7 & 121.2 & 22.6 \\
    \midrule
    DistilDualEnc & - & - & - & - & - & - & 74.16 & 74.30 & 68.05 & - & - & - & - & - \\
    ViLT & 61.5 & 86.3 & 92.7 & 42.7 & 72.9 & 83.1 & 75.7 & 76.1 & 71.3 & - & - & - & - & - \\
    MiniVLM & 58.8 & 85.1 & 91.7 & 45.0 & 74.1 & 84.0 & 73.71 & 73.93 & 69.1 &  69.4 & 35.6 & 28.6 & 119.8 & 21.6 \\ 
    DistillVLM & 58.3 & 84.1 & 91.3 & 43.9 & 73.7 & 83.3 & - & - & 69.8 &  69.6 & 35.6 & 28.7 & 120.8 & 22.1 \\ 
    X-VLM$_\text{small}$ & 74.5 & 92.3 & 96.0 & 56.1 & 81.6 & 88.7 & 79.34 & 79.26 & 73.7 & 73.93 &  37.2 & 29.4 & 123.4 & 22.4 \\
    \method & \bf 78.7 & \bf 94.5 & \bf 97.5 & \bf 60.6 & \bf 84.4 & \bf 90.5 & \bf 81.83 & \bf 81.72 & \bf 76.2 & \bf 76.28 & \bf 38.1 & \bf 30.1 & \bf 127.3 & \bf 23.1 \\
    \bottomrule
    \end{tabular}
    }
    \caption{Main results
 on various downstream vision-language tasks. The top group are teacher models and the 98\% performance of them. The bottom group contains previous efficient VLMs and the X-VLM$_\text{small}$ baseline.
}
\vspace{-2mm}    
\label{tab:main_tbl}
\end{table*}

\subsection{Datasets and Tasks}
\paragraph{Pre-training datasets} We construct our pre-training dataset following \cite{zeng2021multi} 4M-setting using two in-domain datasets, COCO~\cite{lin2014microsoft} and Visual Genome (VG)~\cite{krishna2016visual}, and two out-of-domain datasets, SBU Captions~\cite{ordonez2011im2text} and Conceptual Captions (CC)~\cite{sharma2018conceptual}. Note that we have cleaned the pre-training datasets to avoid data leaks since downstream V+L tasks have overlaps in images with COCO and Visual Genome. The statistics of our pre-training dataset are presented in the Appendix~\ref{app:pretrain_datasets}
\footnotetext{In practice, the text encoder can be run in parallel with image encoder while being much faster. Therefore, the inference time of text encoders does not actually contribute to the overall actual inference time of the model.} .

\noindent \textbf{Image-Text Retrieval} There are two subtasks: text retrieval (TR) and image retrieval (IR). We evaluate X-VLM on MSCOCO datasets. We adopt the widely used Karpathy split~\cite{karpathy2015deep} datasets. Following ALBEF and X-VLM, we optimize $\mathcal{L}_\mathrm{ITC}$ and $\mathcal{L}_\mathrm{ITM}$ and fine-tune the model for 10 epochs. During inference, we first compute $s(I,T)$ for all images and texts, and then take the top-$k$ candidates and calculate $\Vec{p}^\textrm{match}(I,T)$ for ranking. $k$ is set to 256 for MSCOCO following~\citet{zeng2021multi}.

\noindent \textbf{Visual Question Answering} (VQA 2.0)~\citep{goyal2017making} It requires the model to predict an answer given an image and a question. Following ALBEF and X-VLM, we use a three-layer Transformer decoder initialized by the cross-modal encoder of \method to generate answers based on the outputs of the cross-modal encoder. We fine-tune the model for 10 epochs. During inference, we constrain the decoder to only generate from the 3,129 candidate answers following \citet{zeng2021multi,li2021align}.

\noindent \textbf{Natural Language for Visual Reasoning} (NLVR2~\citep{suhr2018corpus}) The task prescribe the model to predict whether a text describes the relations between two images. Following ALBEF and X-VLM, we extend the cross-modal encoder to enable reasoning over two images and performs a domain pre-training step for two epochs. We then fine-tune the model for 10 epochs.

\noindent \textbf{Image Captioning} The task requires a model to generate textual descriptions of input images. We evaluate X-VLM on the COCO Captioning dataset~\citep{chen2015microsoft}. We report BLEU-4~\citep{papineni2002bleu}, METEOR~\citep{denkowski2014meteor}, SPICE~\citep{anderson2016spice} and CIDEr~\citep{vedantam2015cider} scores on the Karparthy test split. Following \citet{zeng2021multi}, we simply adapt \method to a multi-modal decoder for caption generation. We train \method with language modeling loss for two epoch on 4M data. Then, we fine-tune it on the COCO Captioning dataset for 10 epochs.

\subsection{Experiment Setup}

\noindent \textbf{Teacher Models}
We initialized the teacher X-VLM model by a pre-trained CLIP ViT\citep{radford2021learning} and a pre-trained BERT.
We pre-train the X-VLM on 4 million image-text pairs for 20w steps. Then we fine-tune the teacher model on downstream tasks following \citet{zeng2021multi}.   

\noindent \textbf{Pre-training} We pre-train \method on the aforementioned 4 million image-text pairs for 40w steps with 16$\times$ V100 32G GPU. We adopt AdamW\citep{loshchilov2018decoupled} optimizer and set the learning rate and weight decay as 1e-4 and 0.01 respectively. The batch size is set to 1024.

\noindent \textbf{Fine-tuning} We combine the modal-adaptive pruning algorithm with knowledge distillation from the fine-tuned teacher models. We set pruning sparsity at 25\%. Other fine-tuning hyper-parameters are presented in the Appendix~\ref{app:hyper-parameters}.

\begin{table*}[!]
    \centering
    \resizebox{\textwidth}{!}{
    \begin{tabular}{l|ccc|ccc|cc|c|cccc}
    \toprule
        Method & \multicolumn{3}{c|}{ITR-TR} & \multicolumn{3}{c|}{ITR-IR} &\multicolumn{2}{c|}{NLVR2} & VQA &\multicolumn{4}{c}{COCO-Caption}\\
         & R@1 & R@5 & R@10 &  R@1 & R@5 & R@10 & val & test & test-dev &  B@4 & M & C & S\\
         \midrule
         \multicolumn{14}{c}{\textit{Ablation Study Results on Pre-train Distilltion Objectives}} \\
         \midrule
    X-VLM$_\text{small}$ & 73.0 & 91.8 & 96.0 & 55.3 & 81.1 & 88.6 & 78.68 & 78.39 & 73.39 & 35.7 & 29.0 & 117.9 & 21.8 \\
     + Logits & 76.6 & 93.4 & 96.8 & 58.7 & 82.9 & 89.4 & 81.16 & 80.97 & 74.91 & 36.4 & 29.5 & 121.5 & 22.2 \\
    + Hidden & 76.7 & 93.6 & 96.8 & 59.1 & 83.0 & 89.7 & 80.74 & 81.13& 75.12& 36.9 & 29.8 & 126.2 & 22.9  \\
    + Attn & 76.5 & 94.1 & 97.0 & 59.0 & 83.0 & 89.6 & 81.06 & 81.01 & 75.22 & 37.9 & 29.8 & 126.2 & 22.9\\
    
             \midrule
         \multicolumn{14}{c}{\textit{Ablation Study Results on Fine-tuning Objectives}} \\
         \midrule
    \method & 78.7 & 94.5 & 97.5 & 60.6 & 84.4 & 90.5 & 81.83 & 81.72 & 76.2  & 38.1 & 30.1 & 127.3 & 23.1 \\
    - KD only & 78.2 & 94.4 & 97.2 & 60.4 & 84.2 & 90.5 & 82.73 & 81.92 & 76.48 & 38.2 & 30.1 & 127.7 & 23.1\\
    - Pruning only & 77.9 & 94.3 & 97.3 & 59.7 & 83.8 & 90.1 & 80.71 & 80.47 & 74.87 & 6.9 & 10.9 & 8.2 & 3.5\\
    - Fine-tune only & 77.5 & 94.2 & 97.4 & 59.2 & 83.5 & 89.9 & 81.56 & 81.47 & 75.65 & 37.7 & 29.9 & 126.8 & 22.9 \\

    \bottomrule
    \end{tabular}
    }
    \caption{Ablation study results. The top group shows the effects of gradually adding different distilled knowledge at pre-training stage. We take checkpoints at 10w training steps for evaluation. The bottom group presents ablation experiments of pruning and knowledge distillation at fine-tuning stage.
}
\label{tab:ablation}
\end{table*}

\subsection{Experimental Results}

\subsubsection{Main Results}
We present the main results in
Table~\ref{tab:main_tbl}.
The top group of models denotes the base-size VLMs used as the teacher model for different compact VLMs. We also list the 98\% performance of these models for better comparison. Specifically, X-VLM$_\text{clip}$ \footnote{We adopted the first version of XVLM model as teacher instead of the latest one that using Swin-Transformer as its vision encoder because the model architecture of Swin-Transformer maskes the general distillation more difficult.} is the  teacher of \method while OSCAR$_\text{B}$ is the teacher of DistillVLM. In the bottom group, we compare \method with other efficient vision-language models as well as the X-VLM$_\text{small}$ baseline. We can see that \method substantially outperforms all compared models by a large margin despite DistilVLM and MiniVLM are trained with 7 million image-text pairs while \method is only trained with 4 million image-text pairs. Specifically, \method achieves a R@1 of 78.7\% and 60.6\% on Image Retrieval and Text Retrieval respectively, accounting for a large absolute improvement of of 17.2\% and 15.6\% compared to the previous compact SoTA VLMs. We also achieve 81.83\% and 81.72\% accuracy on validation set and test-P set of NLVR2, respectively, surpassing prior efficient VLMs by a large margin. Similar observation can also be found on VQA 2.0 and COCO Captioning, where \method achieves 76.2\% accuracy and 76.28 on test-dev set and test-std set, and 127.3 CIDEr score, respectively. \method also consistently outperforms X-VLM$_\text{small}$ by a large margin on all datasets despite being more compact and efficient, demonstrating the effectiveness of the proposed \textit{distilling then pruning framework}. Moreover, we find that \method surpasses 98\% performance of the teacher model on most datasets. In contrast, DistillVLM underperforms the 98\%  OSCAR$_\text{B}$ baseline by a large margin. Actually, \method recovers 98.4\% performance of X-VLM$_\text{clip}$ on average, while DistilVLM only retains 89.3\% performance of OSCAR$_\text{B}$ on average. This further confirms the effectiveness of our method.

\subsubsection{Ablation Study}
We also conduct a series of ablation study to better understand the effectiveness of \method.

\noindent \textbf{Impact of Knowledge Distillation}
We first investigate the impact of different distillation objectives by starting with a small-size X-VLM model pre-trained with its original objectives only. We then gradually add logits distillation, hidden states distillation and attention distillation. The results are shown in the top group of Table \ref{tab:ablation}. We find that adding each component improves the overall performance, demonstrating the effectiveness of combing these components for pre-train distillation.

\noindent \textbf{Impact of Fine-tuning Objectives}
We then study the effect of modal-adaptive pruning and knowledge distillation in the fine-tuning stage. The results are shown in Table \ref{tab:ablation}. First, by comparing the results of \method and that in Table \ref{tab:observation}, we can see that modal-adaptive pruning with learned sparsity for encoders of each modality substantially outperforms manually tuned sparsity. We also find that \method performs similarly to the KD-only variant. These results confirm the effectiveness of modal-adaptive pruning. We also find that pruning without distillation results in worse results, demonstrating the necessity of knowledge distillation during fine-tuning. Finally, we can see that simply fine-tuning the compact task-agnostic pre-trained \method performs not as well. However, it still outperforms existing baselines by a very large margin. This shows that \method can also be used as a good compact task-agnostic VLM.

\begin{figure}
\centering
\begin{minipage}{.25\textwidth}
  \centering
  \includegraphics[width=\textwidth]{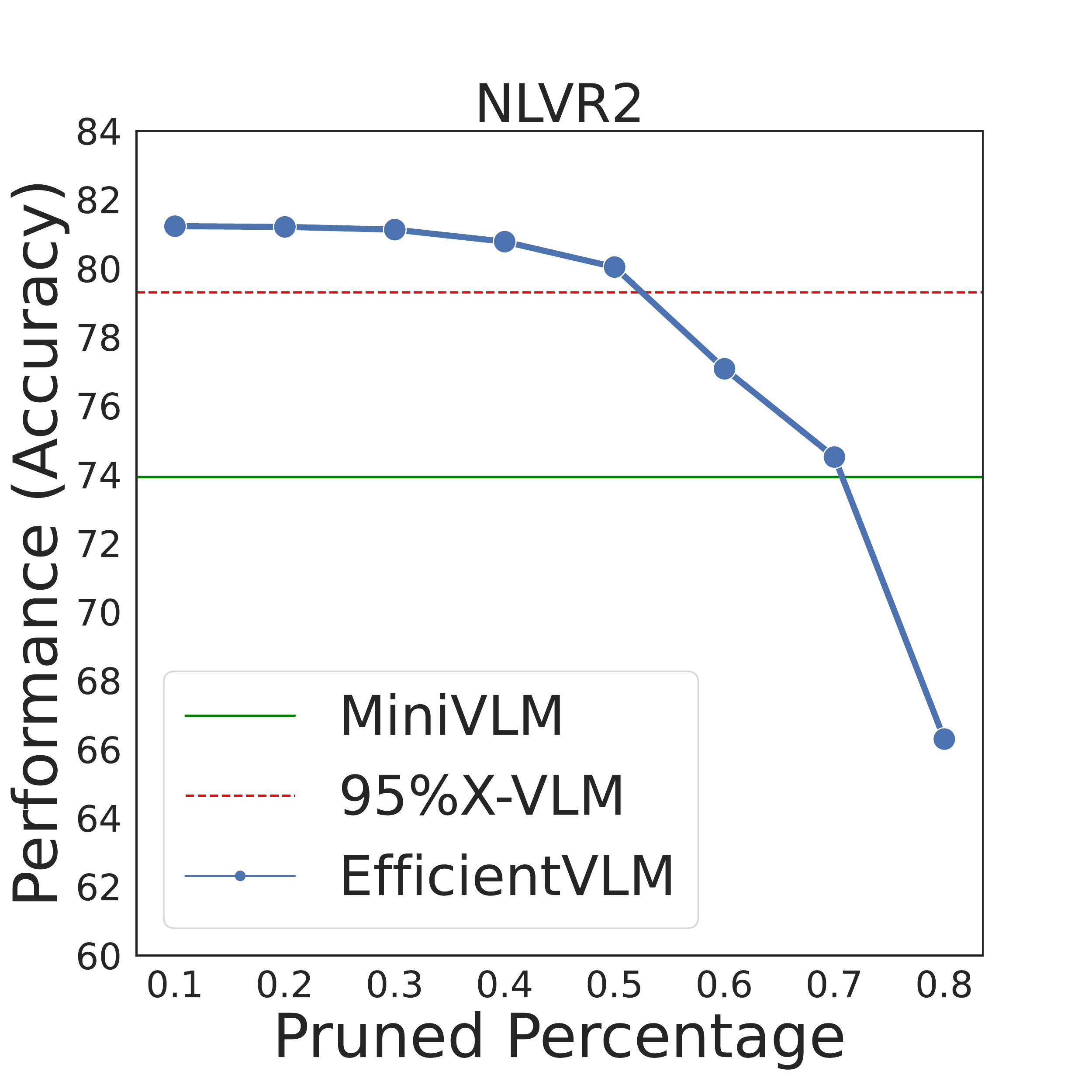}
\end{minipage}%
\begin{minipage}{.25\textwidth}
  \centering
  \includegraphics[width=\textwidth]{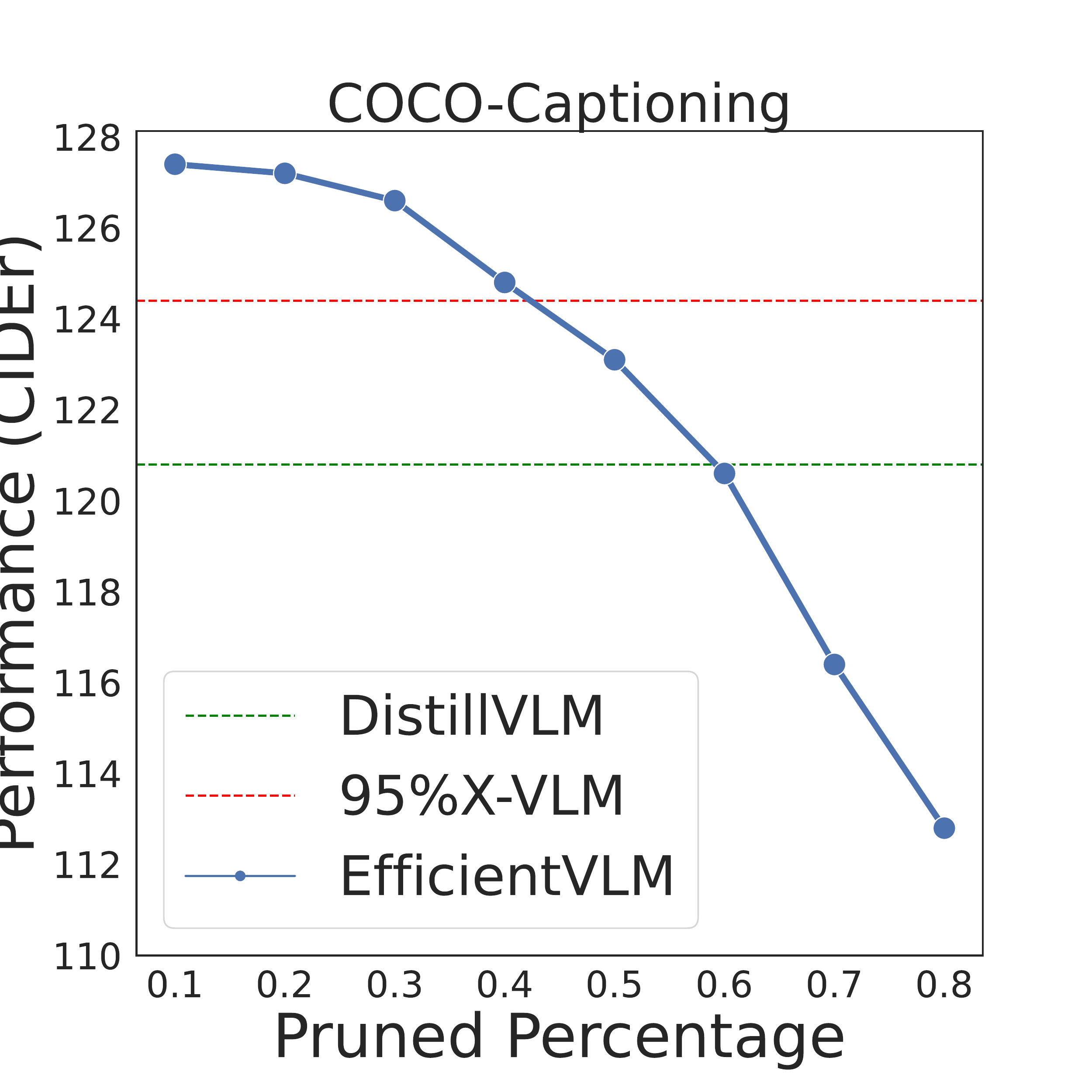}
\end{minipage}
\caption{Ablation study results with different sparsity ranging from 10\% to 80\% on NLVR2 and COCO Captioning datasets.}
\vspace{-0.4cm}
\label{fig:spar_ablation}
\end{figure}

\noindent \textbf{Impact of Pruning Sparsity}
We also investigate the performance of our modal-adaptive pruning methods with different target sparsity ranging from 10\% to 80\%. 
The results are shown in Figure~\ref{fig:spar_ablation}. We can see that \method retains over 95\% performance of the teacher model with a sparsity of 50\% and 40\% on NLVR2 and COCO Captioning, respectively. \method also outperforms previous best results of compact VLMs with a sparsity up to 70\% and 60\% on these tasks. This shows \method also performs well with larger sparsity.

%% file: sections/4_conclusion.tex
\section{Conclusion}
We introduce \method, a fast and accurate vision-language model trained with a distilling then pruning framework. Empirical results show that \method retains 98.4\% performance of the base-size teacher model while only preserving 44.3\% parameters and achieving a speed-up ratio of $2.2\times$. \method also achieves a large absolute improvement over previous efficient VLMs such as DistilVLM and MiniVLM, demonstrating a large potential towards lightweight VLMs.

%% file: sections/5_appendix.tex
\label{sec:appendix}

\section{Differentiable  $L_0$-Norm Regularization}
\label{app:optimization}
 The formulation of Equation~\ref{eq:pruning formulation} is still hard for gradient-based optimization by the discrete nature of masks, but the expectation provides some guidance for empirically effective relaxations. Following prior work\citep{louizos2017learning,wang2019structured,guo2020parameter}, we apply Hard-Concrete distribution~\citep{maddison2017concrete} to relax $\mask$ into continuous space $[0, 1]^d$. 
 Specifically, $\mask$ is now defined to be a deterministic and (sub)differentiable function of a sample $\mathbf{u}$ from a uniform distribution,\vspace{-2mm}
\begin{align*}
& \mathbf{u} \sim U(0,1) \\
& \ \mathbf{s} = \text{sigmoid}(\log \mathbf{u} - \log(1-\mathbf{u}) + \bm{\alpha}) \\
& \ \bar{\mathbf{s}} = \mathbf{s} \times (r - l) + l \\
& \ \mask = \min(1, \max(0, \bar{\mathbf{s}}))
\end{align*}
Here $l < 0$ and $r > 1$ are two constants used to stretch $\mathbf{s}$ into the interval $(l, r)^d$ before it is clamped to $[0, 1]^d$ with the $\min(\mathbf{1}, \max(\mathbf{0}, \cdot))$ operation. In this case we have a differentiable closed-form expression for the expected $L_0$-norm,
\begin{align}
\mathbbm{E}\left[ \|\tilde{\param}\|_0 \right] &= \sum_{j=1}^n \mathbbm{E}\left[ z_j > 0\right] \nonumber \\
&= \sum_{j=1}^n \text{sigmoid}\left(\alpha_j - \log\frac{-l}{r}\right) 
\end{align}

To better control the expected sparsity of the student model, we follow \citet{wang2019structured} to replace the vanilla $l_0$ objective with a Lagrangian multiplier.
Let $t$ be the target model size and $s(\bm{\alpha})$ be the constrained model size determined by the Hard Concrete parameter $\bm{\alpha}$. 

The Lagrangian method imposes an equality constraint $s(\bm{\alpha})=t$ by introducing a violation penalty,
\begin{align*}
\mathcal{L}_\text{Lgr} = \lambda_1 \cdot (s(\bm{\alpha})-t) + \lambda_2 \cdot (s(\bm{\alpha})-t)^2
\end{align*}
where $\lambda_1, \lambda_2 \in \mathbbm{R}$ are two Lagrangian multipliers that will be jointly updated during training.

\section{Pre-train Datasets}
\label{app:pretrain_datasets}
\begin{table}[ht]

	\label{tab:4Mdata}
    \small
	\centering	
	\begin{tabular}	{l | l |  l | l }
	\toprule
	 Dataset & \# Images  & \# Captions & \# Ann \\
\midrule
	  COCO & 0.11M & 0.55M & 0.45M \\
	  VG & 0.10M & - & 5.7M \\
	  SBU & 0.86M & 0.86M & -\\
	  CC-3M & 2.9M & 2.9M & - \\
	 \bottomrule
	\end{tabular}
		\caption{Statistics of the pre-training datasets.}
\end{table}

\section{Hyperparameters}
The hyperparameters to reproduce fine-tuning results are in Table~\ref{tab:hyper-parameters}.Tasks with $^{*}$ need two-stage fine-tuning.
\label{app:hyper-parameters}
\begin{table}[ht]

	\label{tab:FT-hyper-parameter}
    \small
	\begin{tabular}	{l | l |  l | l }
	\toprule
	 Tasks & Learning Rate & Batch Size & Epoch \\
\midrule
	  ITR-COCO & 3e-5 & 384 & 10 \\
	  NLVR$^{*}$ & 3e-5 & 80 & 10 \\
	  Captioning$^{*}$ & 1e-5 & 256 & 5\\
	  VQA & 5e-5 & 192 & 10 \\
	 \bottomrule
	\end{tabular}
		\caption{Hyper-parameters for fine-tuning on downstream tasks.}
\label{tab:hyper-parameters}
\end{table}